\documentclass[letterpaper]{sig-alternate-2013}

\permission{\copyright 2017 International World Wide Web Conference Committee (IW3C2), 
published under Creative Commons CC BY 4.0 License.}
\conferenceinfo{WWW 2017,}{April 3--7, 2017, Perth, Australia.}
\copyrightetc{ACM \the\acmcopyr}
\crdata{978-1-4503-4913-0/17/04. \\
http://dx.doi.org/10.1145/3038912.3052591 \\
\includegraphics{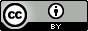}
}

\usepackage{booktabs} % For formal tables
\usepackage{graphicx}
\usepackage{url}
\usepackage{mathptmx}
\usepackage{amssymb}
\usepackage{amsmath}
\usepackage{hyphenat}
\usepackage{xcolor}
\usepackage{textcomp}
\usepackage{enumitem}
\usepackage{tabularx}
\usepackage{mathtools}
\usepackage{bbm}
\usepackage{subfig}
\usepackage{appendix}
\usepackage{cmm-greek}

\newcommand{\hide}[1]{}

\newcommand{\xhdr}[1]{\vspace{1.7mm}\noindent{{\bf #1.}}}

\newcommand{\Secref}[1]{Sec.~\ref{#1}}
\newcommand{\secref}[1]{Sec.~\ref{#1}}

\newcommand{\Tabref}[1]{Table~\ref{#1}}
\newcommand{\tabref}[1]{Table~\ref{#1}}
\newcommand{\Figref}[1]{Fig.~\ref{#1}}
\newcommand{\figref}[1]{Fig.~\ref{#1}}
\newcommand{\question}[1]{\noindent\textit{#1} \\}

\newcommand{\figwidth}{\columnwidth} %0.9
\newcommand{\tabwidth}{0.90\columnwidth}

\DeclareMathAlphabet{\mathcal}{OMS}{cmsy}{m}{n}

\begin{document}%

\title{Ex Machina: Personal Attacks Seen at Scale}

\numberofauthors{3}
\author{
\alignauthor
\hspace{-0.8cm}
Ellery Wulczyn 
% Lucas: I think this is fair; but lets discuss in person before we add explicit quantity of attribution expressions.
\thanks{Equal contribution.}
\\
\hspace{-0.8cm}
       \affaddr{Wikimedia Foundation}\\
\hspace{-0.8cm}
       ellery@wikimedia.org
\alignauthor
\hspace{-2cm}
Nithum Thain
\footnotemark[1]
\\
\hspace{-2cm}
       \affaddr{Jigsaw}\\
\hspace{-2cm}
       nthain@google.com
\alignauthor
\hspace{-3.5cm}
Lucas Dixon \\
\hspace{-3.5cm}
       \affaddr{Jigsaw}\\
\hspace{-3.5cm}
       ldixon@google.com  
}

\maketitle

\begin{abstract}

The damage personal attacks cause to online discourse motivates many platforms to try to curb the phenomenon. However, understanding the prevalence and impact of personal attacks in online platforms at scale remains surprisingly difficult. The contribution of this paper is to develop and illustrate a method that combines crowdsourcing and machine learning to analyze personal attacks at scale. We show an evaluation method for a classifier in terms of the aggregated number of crowd-workers it can approximate. We apply our methodology to English Wikipedia, generating a corpus of over 100k high quality human-labeled comments and 63M machine-labeled ones from a classifier that is as good as the aggregate of 3 crowd-workers, as measured by the area under the ROC curve and Spearman correlation. Using this corpus of machine-labeled scores, our methodology allows us to explore some of the open questions about the nature of online personal attacks. This reveals that the majority of personal attacks on Wikipedia are not the result of a few malicious users, nor primarily the consequence of allowing anonymous contributions from unregistered users.

%This results in several interesting findings: only a small fraction of attacks are currently flagged; while anonymously contributed comments are 6x more likely to be an attack, they contribute less than half of the attacks; and, although the accounts belonging to users with few contributions account for a significant proportion of attacks, users who contribute frequently are also responsible for a significant proportion.

%TODO: consider including 6x more likely + small fraction are moderated.
% \nithum{Alternative last sentences: }\\

% This analysis reveals that personal attacks are not concentrated in the hands of a few highly toxic users on Wikipedia, but rather diffused among many, with a disproportionately high number of attacks coming from anonymous and low activity users. We also show that while attacks tend to cluster and potentially provoke further attacks, only a small fraction of attacks on Wikipedia lead to moderator intervention.
\end{abstract}

\section{Introduction}
\label{sec:intro}
%% !TEX root = 0000_main.tex

% Harassment is a problem; guidelines + policies are the current solution.
With the rise of social media platforms, online discussion has become integral to people's experience of the internet. Unfortunately, online discussion is also an avenue for abuse. A 2014 Pew Report highlights that 73\% of adult internet users have seen someone harassed online, and 40\% have personally experienced it~\cite{duggan14}.
Platforms combat this with policies concerning such behavior. For example Wikipedia has a policy of ``Do not make personal attacks anywhere in Wikipedia''\cite{wp:npa} and notes that attacks may be removed and the users who wrote them blocked.\footnote{This study uses data from English Wikipedia, which for brevity we will simply refer to as Wikipedia.}

% Making good guidelines+polices is difficult, human evaluation is not enough.
The challenge of creating effective policies to identify and appropriately respond to harassment is compounded by the difficulty of studying the phenomena at scale. Typical annotation efforts of abusive language, such as that of Warner and Hirschberg~\cite{warner2012detecting}, involve labeling thousands of comments, however platforms often have many orders of magnitude more; Wikipedia for instance has 63M English talk page comments. Even using crowd-workers, getting human-annotations for a large corpus is prohibitively expensive and time consuming.
%Selecting subsets is difficult because there are an exponential number of ways of selecting a subset from the larger population. Ideally we would be able to annotate the complete corpus.

% Primary contribution: methodology + application to WP.
The primary contribution of this paper is a methodology for quantitative, large-scale, longitudinal analysis of a large corpus of online comments. Our analysis is applicable to properties of comments that can be labeled by crowd-workers with high levels of inter-annotator agreement. We apply our methodology to personal attacks on Wikipedia, inspired by calls from the community for research to understand and reduce the level of \textit{toxic discussions}~\cite{wishlist, consultation}, and by the clear policy Wikipedia has on personal attacks~\cite{wp:npa}.
% The site is curated by over 29 million \cite{wp:num_editors} volunteer contributors and is of the world's most popular online resources \cite{alexa}. Although discussions on Wikipedia are not immediately surfaced to readers of the encyclopedia, every page has an accompanying "talk page", where editors discuss and coordinate their work on compiling and curating the world's knowledge.

% Developing a meaningful classifier for large scale analysis; it's quality is good. 
We start by crowdsourcing a small fraction of the corpus, labeling each comment according to whether it is a personal attack or not. We use this data to train a simple machine learning classifier and experiment with features and labeling methods. The machine learning methods are not novel, but their application does validate and extend the findings of Nobata et al.~\cite{nobata2016abusive}: character-level n-grams result in an impressively flexible and performant classifier for a variety of abusive language in English. We additionally note that using the empirical distribution of human-ratings, rather than the majority vote, produces a better classifier, even in terms of the AUC metric.

The classifier is then used to annotate the entire corpus of comments - acting as a surrogate for crowd-workers. To know how meaningful the automated annotations are, we develop an evaluation method for comparing an algorithm to a group of human annotators. We show that our classifier is as good at generating labels as aggregating the judgments of 3 crowd-workers. To enable independent replication of the work in this paper, as well as to support further quantitative research, we have made public our corpus of both human and machine annotations as well as the classifier we trained~\cite{figshare}. 
% over 100k comments rated 10 times each - the largest available human annotated data sets 
% that we know of on this topic.

% We show how to use it for large scale analysis.
We use our classifier's annotations to perform quantitative analysis over the whole corpus of comments. To ensure that our results accurately reflect the real prevalence of personal attacks within different sub-groups of comments, we select a threshold that appropriately balances precision and recall. We also empirically validate that the threshold produces results on subgroups of comments commensurate with the results of crowd-workers. 

This allows us to answer questions that our much smaller sample of crowdsourced annotations alone would struggle to. We illustrate this by showing how to use our method to explore several open questions about the nature of personal attacks on Wikipedia: What is the impact of allowing anonymous contributions, namely those from unregistered users? How do attacks vary with the quantity of a user's contributions? Are attacks concentrated among a few users? When do attacks result in a moderator action? And is there a pattern to the timing of personal attacks? 

%The policy takes a broad approach, providing a non-exhaustive list of examples that includes "racial, sexist, homophobic, ageist, religious, political, ethnic, national, sexual, or other epithets", ad hominem attacks, and threats.
%The full corpus of machine-labeled Wikipedia discussions has been made public \cite{figshare} and enables other researchers and members of the community with basic data science skills to perform a broad array of large-scale quantitative analyses into the nature of personal attacks. In the present work, we include some sample investigations into the following broad research questions: 

% \begin{enumerate}[nolistsep]
% \item How prevalent are personal attacks on Wikipedia?
% \item What kinds of users make personal attacks?
% \item How do moderators react to personal attacks?\\
% \end{enumerate}

The rest of the paper proceeds as follows: \secref{sec:relatedwork} discusses related work on the prevalence, impact, and detection of personal attacks and closely related online behaviors. In \secref{sec:data} we describe our data collection and labeling methodology. \secref{sec:modeling} covers our model-building and evaluation approaches. We describe our analysis of personal attacks in  Wikipedia in \secref{sec:analysis}. We conclude in \secref{sec:conclusion} and outline challenges with our method and possible avenues of future work.

\vspace{0.2cm}

\section{Related work}
\label{sec:relatedwork}
% !TEX root = 0000_main.tex

\noindent\textbf{Definitions, Prevalence and Impact.} One of the challenges in studying negative online behavior is the myriad of forms it can take and the lack of a clear, common definition \cite{ruths16}. While this study focuses on personal attacks, other studies explore different forms of online behavior including hate speech (\cite{gagliardone15}, \cite{kwok13}, \cite{ruths16}, \cite{warner2012detecting}), online harassment (\cite{cheng15antisocial}, \cite{yin09}), and cyberbullying (\cite{pieschl2015beware}, \cite{schrock11}, \cite{tokunaga10}, \cite{willard07}, \cite{xu2013examination}).

\begin{figure}[h]
\centering
\includegraphics[width=\columnwidth]{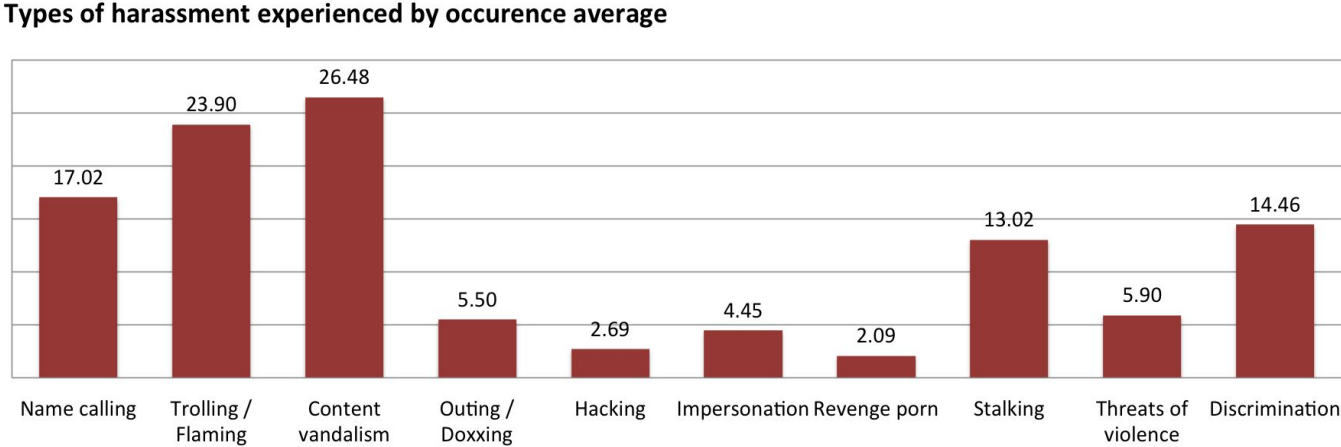}
\caption{
Forms of harassment experienced on Wikimedia \cite{wikisurvey}.
}
\label{fig:wmf_survey}
\end{figure}

Online harassment itself is sometimes further divided into a taxonomy of forms. A recent Pew Research Center study defines online harassment to include being: called offensive names, purposefully embarrassed, stalked, sexually harassed, physically threatened, and harassed in a sustained manner \cite{duggan14}. The Wikimedia Foundation Support and Safety team conducted a similar survey \cite{wikisurvey} using a different taxonomy (see Figure \ref{fig:wmf_survey}).

This toxic behavior has a demonstrated impact on community health both on and off-line. The Wikimedia Foundation found that 54\% of those who had experienced online harassment expressed decreased participation in the project where they experienced the harassment~\cite{wikisurvey}. Online hate speech and cyberbullying are also closely connected to suppressing the expression of others~\cite{sood2012automatic}, physical violence~\cite{wiener1998negligent}, and suicide~\cite{dinakar2011modeling}. \\

\noindent\textbf{Automated Detection.} There have been a number of recent papers on detecting forms of toxic behavior in online discussions. Much of this work builds on existing machine learning approaches in fields like sentiment analysis \cite{pang2008opinion} and spam detection \cite{spirin2012survey}. On the topic of harassment, the earliest work on machine learning based detection is Yin et al.'s 2009 paper \cite{yin09} which used support vector machines on sentiment and context features extracted from the CAW 2.0 dataset \cite{caw2}. In \cite{sood2012automatic}, Sood et al. use the same algorithmic framework to detect personal insults using a dataset labeled via Amazon Mechanical Turk from the Yahoo! Buzz social news site. Dinakar et al. \cite{dinakar2011modeling} decompose the issue of cyberbullying by training separate classifiers for attacks based on sexual orientation, race or intelligence in YouTube comments. Building on these works, Cheng et al. \cite{cheng15antisocial} use random forests and logistic regression techniques to predict which users of the comment sections of several news sites would become banned for antisocial behavior. Most recently, Nobata et al. \cite{nobata2016abusive} extract character n-gram, linguistic, syntactic, and distributional semantic features from a very large corpus of Yahoo! Finance and News comments to detect abusive language. \\

%One of the earliest works on the algorithmic detection of hate speech was that of Warner and Hirschberg \cite{warner2012detecting}, who use support vector machines to classify anti-semitic text in paragraphs from websites and comments in Yahoo! groups. Kwok and Wang \cite{kwok13} extend this approach and use a Naive Bayes classifier to detect anti-black hate speech on Twitter. Ting et al \cite{ting2013approach} combine text features and social network features to detect hate groups on Facebook. Finally, in \cite{ruths16}, Saleem et al use known online hate groups and support groups to generate large scale training data for a range of machine learning algorithms to classify hate speech online. \ellery{commenting this out, since we do not talk about hate speech in the paper}

%\subsection{Data Sets} \label{subsec:datasets}

\noindent\textbf{Data Sets.} A barrier to further algorithmic progress in the detection of toxic behavior is a dearth of large publicly available datasets \cite{ruths16}. To our knowledge, the current open datasets are limited to the Internet Argument Corpus~\cite{walker2012corpus},  
the CAW 2.0 dataset provided by the Fundacion Barcelona Media~\cite{caw2}, the "Hate Speech Twitter Annotations" corpus\cite{waseem2016hateful}, and the "Detecting Insults in Social Commentary" dataset released by Impermium via Kaggle \cite{impermium}. In past work, many researchers have relied on creating their own hand-coded datasets (\cite{kwok13}, \cite{sood2012automatic}, \cite{warner2012detecting}), using crowd-sourced or in-house annotators.  These approaches limit the size of the labeled corpora due to the expense of labeling examples. A few authors have suggested alternative techniques that could be effective in obtaining larger scale datasets. In \cite{ruths16}, Saleem et al. outline some of the limitations of using a small hand-coded dataset and suggest an alternative approach that uses all comments within specific online communities as positive and negative training examples of hate speech. Xiang et al. \cite{xiang2012detecting} use topic modeling approaches along with a small seed set of tweets to produce a training set for detecting offensive tweets containing over 650 million entries. Building on the work of \cite{yin09}, Moore et al. \cite{moore2012anonymity} use a simple rule-based algorithm for automatic labeling of forum posts on which they wish to do further analysis.

%Understanding personal attacks
%cite cheng

%Amplified asking
% cite Salganik, Blumenstock

\section{Crowdsourcing}
\label{sec:data}
% !TEX root = 0000_main.tex

In this section we discuss our approach to identifying personal attacks in a subset of Wikipedia discussion comments via crowdsourcing. The crowdsourcing process involves: \\

\begin{enumerate}[nolistsep]
\item generating a corpus of Wikipedia discussion comments,
\item choosing a question for eliciting human judgments,
\item selecting a subset of the discussion corpus to label,
\item designing a strategy for eliciting reliable labels.\\
\end{enumerate}

To generate a corpus of discussion comments, we processed the public dump of the full history of English Wikipedia as described in Appendix \ref{appendix:corpus}.  
The corpus contains 63M comments from discussions relating to user pages and articles dating from 2004-2015. 

\begin{figure}[h]
\centering
\includegraphics[width=\columnwidth]{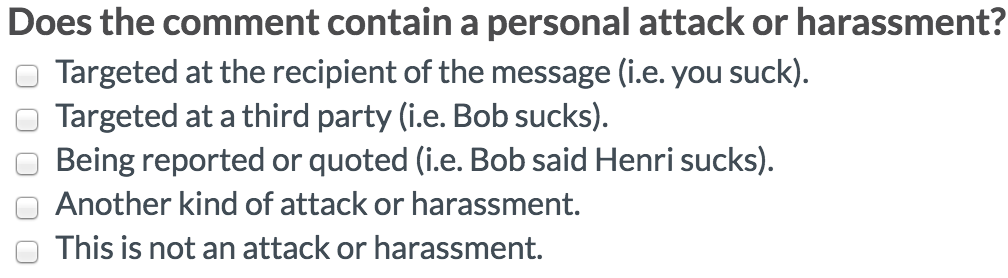}
\caption{The question posed to our Crowdflower annotators.}
\label{fig:crowdflower_question}
\end{figure}

%question
The question we posed to get human judgments on whether a comment contains a personal attack is shown in Figure \ref{fig:crowdflower_question}. In addition to identifying the presence of an attack, we also try to elicit if the attack has a target or whether the comment quotes a previous attack. We do not, however, make use of this additional information in this study. 
%Instead, we consider an annotation to have a value of 0 or "not attack" if the annotator selected the option "This is not an attack or harassment" and a value 1 or "attack" otherwise. 
Before settling on the exact phrasing of the question, we experimented with several variants and chose the one with the highest inter-annotator agreement on a set of 1000 comments.

% data selection

To ensure representativeness, we undertook the standard approach of randomly sampling comments from the full corpus. We will refer to this set of comments as the \textit{random} dataset. Through labeling a random sample, we discovered that the overall prevalence of personal attacks on Wikipedia talk pages is around 1\% (see Section \ref{subsec:threshold}). 

%This means we would need an enormous sample for it to contain a large enough number of personal attacks to build a strong classier.
To allow training of classifiers, we need to create a corpus that contains a sufficient number and variety of examples of personal attacks. In order to obtain these, we enhance our random dataset by also sampling comments made by users who where blocked for violating Wikipedia's policy on personal attacks \cite{wp:npa}. 
In particular, we consider the 5 comments made by these users around every block event. We call this the \textit{blocked} dataset and note that it has a much higher prevalence of attacks (approximately 17\%).

%\nithum{I removed this to simplify. Let me know if you'd like to revert.}
%In order to guide our decision on how what proportions of \textit{random} and \textit{blocked} data to label, we started by labeling 10k comments each type. Then we evaluated a simple bag-of-words text classifier on a test set of \textit{random} data, while using a training data set of fixed size but varying ratios of \textit{random} and \textit{blocked} data. We found that adding \textit{blocked} data increased ROC by up to 5 points and that there where diminishing, though still positive returns after training data with 70\% \textit{blocked} data.

\begin{table}[h]
\centering
\noindent \begin{tabularx}{\tabwidth}{X X X}
\toprule
Sample Type & Annotated \mbox{Comments} & Percentage Attacking \\
\midrule
Random  & 37611  & 0.9 \% \\
Blocked  & 78126  & 16.9 \% \\
\midrule
Total & 115737  & 11.7 \%  \\
\bottomrule
\end{tabularx}
\caption{Summary statistics of labeled data. Each comment was labeled 10 times. Here we define a comment as an attack if the majority of annotators labeled it as such.}
\label{tab:annotated_datasets}
\end{table}

%\begin{table}[h]
%\centering
%\noindent \begin{tabularx}{\tabwidth}{ X X X X}
%  \toprule
%  Namespace & Sample Type & Annotated Comments & Percentage Attacking \\
%  \midrule
%  User  & Random  & 18040  & 1.1 \\
%    & Blocked  & 46705  & 23.9 \\
%  \midrule
%
%  Article  & Random  & 19571  & 0.8 \\
%    & Blocked  & 31421  & 6.6 \\
%  \midrule
%  \midrule
%  Totals &  & 115737  & 11.7  \\
%  \bottomrule
%\end{tabularx}
%\caption{Summary statistics of labeled data. Each comment was labeled 10 times. Here we define a comment as an attack if the majority of annotators labeled it as such.}
%\label{tab:annotated_datasets}
%\end{table}

% crowdsourcing scheme
We labeled our subset of comments using the Crowdflower crowdsourcing platform.\footnote{https://www.crowdflower.com/} Crowdsourcing as a data collection methodology is well studied (\cite{buhrmester2011amazon}, \cite{tetreault2010rethinking}) and has proven effective for constructing corpora for machine learning in various contexts (\cite{cheng15antisocial},
\cite{kwok13}, \cite{plank2014learning}, \cite{sood2012automatic}, \cite{warner2012detecting}).

% TODO: consider clarifying.
As a first step to ensuring data quality, each annotator was required to pass a test of 10 questions. These questions were randomly selected from a set that we devised to contain balanced representation of both attacking and non-attacking comments. Annotators whose accuracy on these test questions fell below a 70\% threshold would be removed from the task. This improved our annotator quality by excluding the worst \textasciitilde 2\% of contributors. Under the Crowdflower system, additional test questions are randomly interspersed with the genuine crowdsourcing task (at a rate of 10\%) in order to maintain response quality throughout the task.

In order to get reliable estimates of whether a comment is a personal attack, each comment was labeled by at least 10 different Crowdflower annotators. This allows us to aggregate judgments from 10 separate people when constructing a single label for each comment. We chose 10 judgments based on experiments in~\secref{subsec:baselines} that showed that aggregating more judgments provided little further improvement. 
%\ellery{how do we chose how many judgments to collect per comment. Commenting out, it's a long story...}
Finally, we applied several data cleaning steps to the Crowdflower annotations. This included removing annotations where the same worker labeled a comment as both an attack and not an attack and removing comments that most workers flagged as not being English.

% evaluation
We evaluated the quality of our crowd-sourcing pipeline by measuring inter-annotator agreement~\cite{krippendorff2004content}. This technique measures whether a set of ``common instructions to different observers of the same set of phenomena, yields the same data within a tolerable margin of error''~\cite{hayes2007answering}. We chose the specific inter-annotator agreement metric of Krippendorf's alpha due to our context, where multiple raters rate overlapping, but disparate sets of comments \cite{krippendorff2004reliability}. Our data achieves a Krippendorf's alpha score of 0.45. This result is in-line with results achieved in other crowdsourced studies of toxic behavior in online communities~\cite{cheng15antisocial}. 

%% This is said already in conclusion. 
% A contribution of our research is the public release of this dataset \cite{figshare}. We believe this is the largest dataset supporting the study of online toxicity to date.

\section{Model Building}
\label{sec:modeling}
% !TEX root = 0000_main.tex
We now use the set of crowdsourced annotations to build a machine learning classifier for identifying personal attacks. 
% Our goal is to use this classifier to effectively amplify the crowdsourcing process to cover a corpus of millions of comments. 
We first discuss the set of machine learning architectures we explored and then describe our evaluation methodology. 
% We show that a relatively simple classifier can perform as well at identifying personal attacks as pooling the judgments of three crowd-workers.

\subsection{Model Building Methodology}
\label{subsec:modeling_methodology}

% models and features
We treat the problem of identifying personal attacks as a binary text classification problem. We rely purely on features extracted from the comment text instead of including features based on the authors' past behavior and the discussion context. This makes it easy for Wikipedia editors and administrators, journalists and other researchers to explore the strengths and weaknesses of the models by simply generating text examples. It also allows the models to be applied beyond the context of Wikipedia. 

In terms of model architectures, we explored logistic regression (LR), and multi-layer perceptrons (MLP). 
% If we are shot on space, we can remove the next sentence.
In future work, we plan to experiment with long short-term memory recurrent neural networks (LSTM) as well. 
For the LR and MLP models we simply use bag-of-words representations based on either word- or character-level n-grams. Past work in the domain of detecting abusive language in online discussion comments, showed that simple n-gram features are more powerful than linguistic and syntactic features, hand-engineered lexicons, and word and paragraph embeddings \cite{nobata2016abusive}.
% For the LSTM models, we experimented with both word- and \ellery{character-level sequences}.

%labels
In all of the model architectures, we have a final softmax layer and use the cross-entropy as our loss function. The cross-entropy function is defined as:

\begin{equation}
H(y, \hat{y}) = -\sum_i y_i \log(\hat{y_i})
\end{equation}

\noindent where $\hat{y}$ is our predicted probability distribution over classes, and $y$ is the true distribution. 

In addition to experimenting with different model architectures, we also experimented with two different ways of synthesizing our 10 human annotations per comment to create training labels. In the traditional classification approach, there is only one true class and so the true distribution, $y$, is represented as a one-hot (OH) vector determined by the majority class in the comment's set of annotations.

For the problem of identifying personal attacks, however, one can argue that there is no single true class. Different people may judge the same comment differently. 
%Furthermore, comments can differ in the level of agreement that a group of people displays in their judgments. 
Unsurprisingly, we see this in the annotation data: most comments do not have a unanimous set of judgments, and the fraction of annotators who think a comment is an attack differs across comments.
%\Figref{fig:agreement} shows a histogram of the fraction of annotators who think a comment is an attack over all comments.

%\begin{figure}[h]
%\centering
%\includegraphics[width=8.5cm]{./figs/agreement.png}
%\caption{Histogram of the fraction of annotators who think a comment is an attack across all comments.}
%\label{fig:agreement}
%\end{figure}

The set of annotations per comment naturally forms an approximate empirical distribution (ED) over opinions of whether the comment is an attack. A comment considered a personal attack by 7 of 10 annotators can thus be given a true label of [0.3, 0.7] instead of [0,1].   
% We hypothesize, that training with the ED labels gives a boost in performance over using the OH labels, since they contain more information.
Using ED labels is motivated by the intuition that comments for which 100\% of annotators think it is an attack are probably different in nature from comments where only 60\% of annotators consider it so. Since the majority class is the same in both cases, the OH labels lose the distinction. Hence, in addition to the OH labels, we also trained each architecture using ED labels.
% The ED based models on the other hand incur a higher loss for giving a predicted probability of being an attack of 40\% in the case where 100\% of annotators thought the comment was an attack than in the case where only 60\% of annotators thought the comment was an attack.

Finally, we should note that the interpretation of a model's scores depends on whether it was trained on ED or OH labels. In the case of a model trained on ED labels, the attack score represents the predicted fraction of annotators who would consider the comment an attack. In the case of a model trained on OH labels, the attack score represents the probability that the majority of annotators would consider the comment an attack.

\subsection{Model Building Evaluation}
\label{subsec:modeling_evaluation}

As discussed above, we considered three major dimensions in the model design space:

\begin{enumerate}[noitemsep]
  \item model architecture (LR, MLP)
  \item n-gram type (word, char)
  \item label type (OH, ED)
\end{enumerate}

In order to evaluate each of the 8 possible modeling strategies we randomly split our set of annotations into train, development and test splits (in a 3:1:1 ratio).
% Consider adding a note on why 3:1:1 (its a bit unusual)
For each model, we performed 15 iterations of random search over a grid of relevant hyper-parameters\cite{random_search}.\footnote{For details on the set of hyper-parameters explored, we refer the reader to the relevant notebook in our code repository: \url{https://github.com/ewulczyn/wiki-detox/blob/master/src/modeling/cv\_ngram\_architectures.ipynb}}
During the model tuning process, each run was trained on the train split and evaluated on the development split. \tabref{tab:metrics} shows two evaluation metrics for each of the 8 tuned models. The standard 2-class area under the receiver operating characteristic curve (AUC) score is computed between the models' predicted probability of being an attack and the majority class label in the set of annotations for each comment. To better evaluate the performance of models trained on ED labels, we also include the Spearman rank correlation between the models' predicted probability of being an attack and the fraction of annotators who considered the comment an attack.

\begin{table}[h]
\centering
\noindent \begin{tabularx}{\tabwidth}{ X X X X X}
  \toprule
  Model Type & N-Gram Type & Label Type & AUC & Spearman \\
  \midrule
  LR  & Word  & OH  & 94.62 &  53.16\\
    &   & ED  & 95.55 &  65.2\\
    & Char  & OH  & 96.18 &  59.20\\
    &   & ED  & 96.24 &  66.68\\
  \midrule
  MLP  & Word  & OH  & 95.25 &  56.11\\
    &   & ED  & 96.15 &  66.33\\
    & Char  & OH  & 95.90 &  58.77\\
    &   & ED  & \textbf{96.59} &  \textbf{68.17}\\
%  \midrule
%  LSTM  & Word  & OH  & 95.79 &  54.17\\
%    &   & ED  & 96.60 &  66.73\\
%    & Char  & OH  & xx &  xx\\
%    &   & ED  & xx &  xx\\

  \bottomrule
\end{tabularx}
\caption{Evaluation metrics of different model architectures trained on the train split and evaluated on the development split. The hyper-parameters of each architecture were tuned using randomized search}
\label{tab:metrics}
\end{table}

Across all model and label types, we see that character n-grams outperform word n-grams, which is consistent with feature importance analysis in \cite{nobata2016abusive}.\footnote{Note we explored word n-grams in the range of 1-2 and characters n-grams in the range 1-5. During the hyper-parameters search, we searched over the same range of values for the number of n-gram features to include.} We suspect this is due to the higher robustness to spelling variations that char-n-grams exhibit, which are very common in online discussions, especially in expletives commonly used in personal attacks. 

Another consistent pattern is the boost in the performance metrics for models trained using ED labels. The large boost in Spearman correlation is somewhat unsurprising because it is a function of the fraction of annotators who consider a comment to be a personal attack. The models trained using OH labels did not receive any supervision on how to estimate this fraction (they only see majority class). The interesting result is that even the AUC scores are consistently higher: this means that using training labels that encode what fraction of people think a comment is an attack helps in predicting what the majority of annotators think. Our results indicate that using ED labels may give a performance boost over the standard OH labels for other machine-learning tasks using multiple crowdsourced labels per training example.

\subsection{Human Baseline Comparison}
\label{subsec:baselines}
%baselines
We developed a classifier for detecting personal attacks in order to score the full history of comments on Wikipedia in a cost and time effective manner. For our purposes, the model is an approximation of the crowdsourcing process. Hence, we want to be able to answer the question: How good of a surrogate is our model for crowdsourced annotations? 

To answer this we will use one group of annotators, call them the prediction-group, to predict what another group of annotators, call them the truth-group, thinks about a comment. We treat the aggregated judgments of the truth-group as ground truth labels. We treat the prediction-group as a model: an ensemble of annotators, who pool their judgments to make predictions. Hence, we will refer to the prediction-group as the \textit{annotator ensemble}. By comparing our machine learning model's predictive power to the predictive power of the annotator ensemble, we can get an estimate of how good of a surrogate our model is for a fixed size annotator ensemble. We will refer to this method of generating baselines as \textit{annotator ensemble baselining}.

To be more specific, let's fix the size of the truth-group at $n_t$ and the size of the prediction-group at $n_p$. Assume we have collected at least $n_t + n_p$ annotations per comment in our corpus. Now, for each comment $c$, we randomly split the full set of annotations for $c$ into two non-overlapping sets, $T_{c}$ for the truth-group and $P_c$ for the prediction-group, of sizes $n_t$ and $n_p$ respectively. We split the set of annotators at the comment level, since the corpus of comments may be so large that not every annotator judges every comment, making a fixed split across all comments impossible in general. We will aggregate annotations in $T_c$ using the function $agg_{true}$ to get a ground truth label $y(c)$ for comment $c$. The choice of $agg_{true}$ depends on the evaluation metric we want to use. For the AUC metric $agg_{true}$ is the OH aggregation function, whereas for Spearman correlation $agg_{true}$ is the ED aggregation function. We will take the average of the annotations in $P_c$ to get a prediction $\hat{y}(c)_{AE}$. We apply our machine learning model, to comment $c$ to get prediction $\hat{y}(c)_{ML}$. Finally, we compute the evaluation metrics of AUC and Spearman correlation over the entire corpus between $y$ and $\hat{y}_{ML}$ and between $y$ and $\hat{y}_{AE}$ to compare the machine learning model to the annotator ensemble. The comparison of these scores, tells us how good our model is compared to an ensemble of annotators of size $n_p$ at predicting labels generated by pooling $n_t$ annotations. 

Note that, for each question, the annotators are randomly split into a prediction-group and a truth-group. As a result, there is some variability in the evaluation metrics stemming from this assignment step. By running the entire process several times, we can estimate this variability and average the evaluation metric results from each run to get a more stable estimate. 

We applied our \textit{annotator ensemble baselining} method to a set of 8,000 comments from the test split and had each comment labeled 20 times. We will refer to this special subset of comments as the baseline split. Out of the baseline split, 4000 comments come from our \textit{random} dataset and 4000 come from our \textit{blocked} dataset. We fix $n_t$, the number of annotations used to generate labels, at 10, since this is the number of annotations per comment used in training the model. \Tabref{tab:baselines} shows AUC scores and Spearman correlations for the aggregate prediction of the prediction-group as we vary its size $n_p$ from 1 to 10. The final line of the table also reports the values for the best LR model architecture from \tabref{tab:metrics}.\footnote{Note that the reported performance of our model is slightly different in \tabref{tab:metrics} than in \tabref{tab:baselines}, since in the former table the model is evaluated on the dev split, while in the latter table it is evaluated in the baseline split and ratios of random to blocked comments differ across the two splits.} The reported mean scores and standard errors are the result of running the entire process 25 times.
% TODO: it's unclear at this point, if 'the process' includes getting the 8000 comments comments.

For the annotator ensemble, both the AUC scores and Spearman correlations increase with diminishing returns as the size of the ensemble increases. On both of our metrics, our model outperforms an annotator ensemble of size $n_p = 3$. Thus, by these two metrics, running our model over the full history of comments in Wikipedia is as good as having each comment labeled by 3 annotators. \\

\begin{table}[h]
\centering
\noindent \begin{tabularx}{\tabwidth}{ X X X}
\toprule
$n_p$ & AUC & Spearman \\
\midrule
1    &   88.54  \hspace{0.1 cm}  (0.42)   &  53.58 \hspace{0.1 cm} (0.79) \\     
%2    &   95.86  \hspace{0.1 cm}  (0.34)   &  61.45 \hspace{0.1 cm} (0.53) \\     
3    &   95.49  \hspace{0.1 cm}  (0.31)   &  64.75 \hspace{0.1 cm} (0.44) \\     
%4    &   97.31  \hspace{0.1 cm}  (0.24)   &  66.81 \hspace{0.1 cm} (0.50) \\     
5    &   97.13  \hspace{0.1 cm}  (0.23)   &  68.27 \hspace{0.1 cm} (0.46) \\     
%6    &   98.05  \hspace{0.1 cm}  (0.18)   &  69.21 \hspace{0.1 cm} (0.53) \\     
7    &   97.81  \hspace{0.1 cm}  (0.15)   &  69.86 \hspace{0.1 cm} (0.60) \\     
%8    &   98.37  \hspace{0.1 cm}  (0.11)   &  70.36 \hspace{0.1 cm} (0.40) \\     
9    &   98.24  \hspace{0.1 cm}  (0.14)   &  70.97 \hspace{0.1 cm} (0.44) \\     
10   &   98.53  \hspace{0.1 cm}  (0.12)   &  71.11 \hspace{0.1 cm} (0.36) \\     

\midrule
Model: & 97.19 \hspace{0.1 cm} (0.14) & 66.02 \hspace{0.1 cm}  (0.44) \\
\bottomrule
\end{tabularx}
\caption{Mean evaluation metrics (and standard errors) on the baseline split, fixing the truth-group size $n_t$ at 10 and varying the prediction-group size $n_p$. }
\label{tab:baselines}
\end{table}

%\nithum{Should we do a comparison with other papers that do machine learning approaches to show that we are state of the art? In particular, I'm thinking about the Yahoo paper from WWW 2016.} \\

%\begin{figure}[t]
%\centering
%\includegraphics[width=8.5cm]{./figs/precision_recall_curve.png}
%\caption{Precision and recall curves of the char-ngram MLP using ED labels. Computed on the test split.}
%\label{fig:attack_vs_anon}
%\end{figure}

\section{Analysis}
\label{sec:analysis}
% !TEX root = 0000_main.tex

Using the best personal attack classifier from \secref{subsec:modeling_evaluation}, we obtain a full corpus of machine-labeled discussions in Wikipedia. In this section, we use the fully annotated corpus to better understand the prevalence and nature of attacks in Wikipedia. For the following analyses, we focus on comments made in 2015 and exclude administrative comments and comments generated by bots as described in Appendix A.

\subsection{Choosing a Threshold}
\label{subsec:threshold}
Given a comment, our classifier outputs a continuous score in the interval $[0, 1]$. To get a discrete label from this score, we pick a threshold $t$ and let comments with a score above $t$ have label 1, indicating an attack. Using discrete labels makes it possible to identify individual comments that are predicted to contain personal attacks and estimate the fraction of attacks within a set of comments. 
%Although it may be desirable to use different thresholds in different circumstances, we pick a single threshold for all of our subsequent analyses in order to simplify the discussion. In particular, 

To choose a threshold, we pick the point that strikes a balance between precision and recall on $random$ evaluation data. A key property of this threshold for the purpose of using machine-generated labels for analysis, is that false positives are offset by false negatives. As a result, the fraction of comments that are labeled as attacks by the classifier is the same as the fraction of comments that are labeled as attacks by the human annotators. Hence, we refer to this threshold as the \textit{equal-error threshold}.\footnote{This threshold also maximizes the F1 score, since our precision and recall are monotonic functions of the decision threshold.} 

\begin{figure}[h]
\centering

\subfloat[]
{\includegraphics[width=0.14\textwidth]{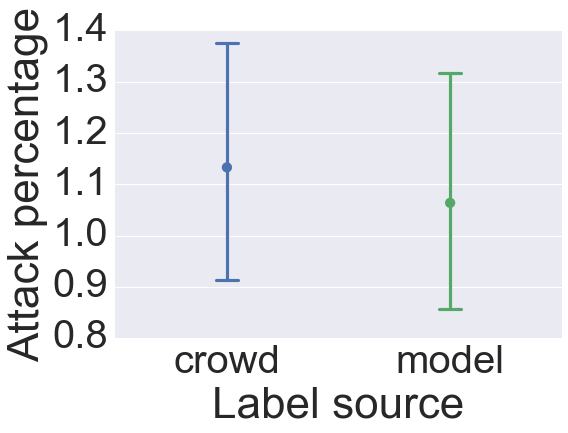}\label{fig:sanity_check}}
\hfill
  \subfloat[]
{\includegraphics[width=0.14\textwidth]{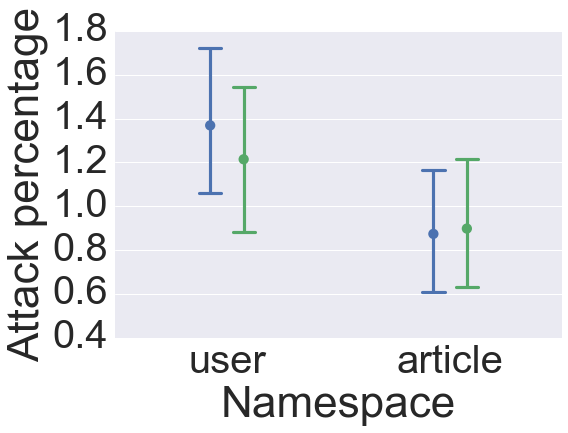}\label{fig:sanity_check_ns}}
\hfill
\subfloat[]
{\includegraphics[width=0.14\textwidth]{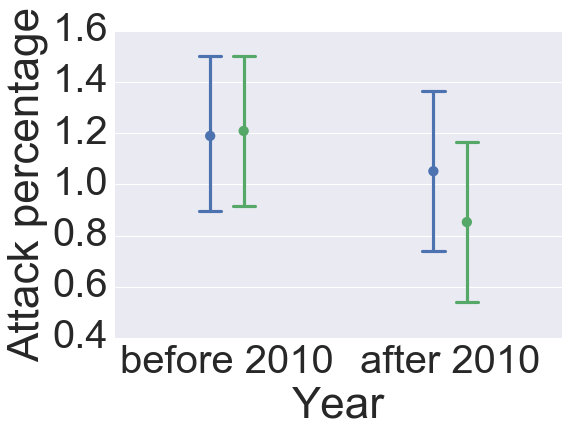}\label{fig:sanity_check_year}}

\subfloat[]
{\includegraphics[width=0.14\textwidth]{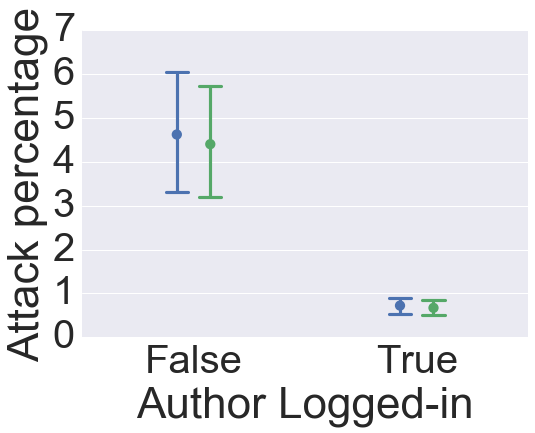}\label{fig:sanity_check_anon}}
\hfill
\subfloat[]
{\includegraphics[width=0.14\textwidth]{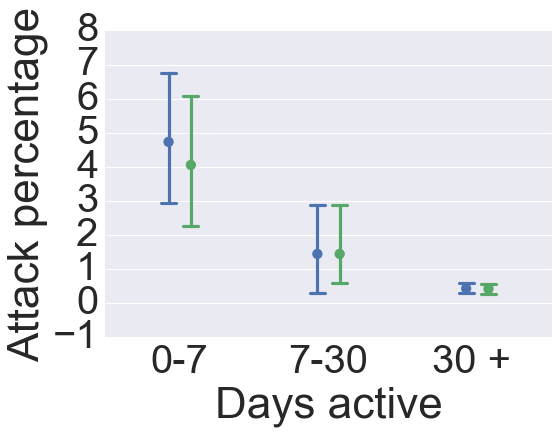}\label{fig:sanity_check_activity}}
\hfill
\subfloat[]
{\includegraphics[width=0.14\textwidth]{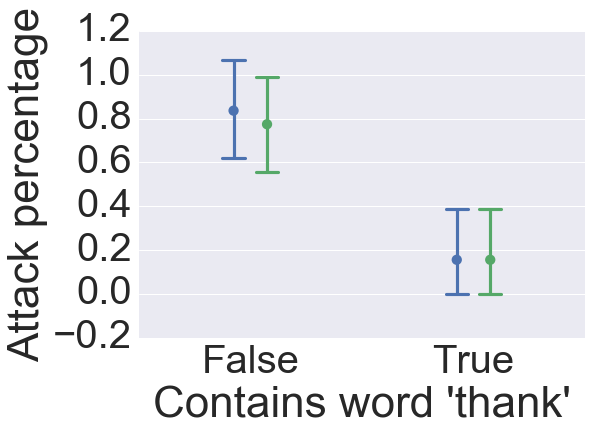}\label{fig:sanity_check_thank}}

 \caption{ Bootstrapped confidence intervals for the fraction of attacks within the test set broken down by (b) discussion namespace, (c) year, (d) logged in status of the author, (e) number of days author made an edit as of 2015, (f) whether the comment contains the word "thank". Blue intervals come from ground-truth human labels, green intervals come from machine generated labels.}

 \end{figure}

To see how well this property generalizes to new data, we used the \textit{equal-error threshold} on the development set and used it to get model-generated labels for the test set. \figref{fig:sanity_check} shows that the estimated rate of attacks computed using model-generated labels lies within a 95\% confidence interval for the rate of attacks computed from crowd-generated labels.

Even though the thresholded model-scores give good estimates of the rate of attacks over a random sample of comments, it is not given that they also give accurate estimates when partitioning comments into different groups. To provide empirical evidence that the thresholded scores give accurate estimates when subdividing the dataset into groups that will be important for later analysis, we shows various splits of the dataset in \figref{fig:sanity_checks}. We split on the year the comment was posted, by whether the author was logged-in, by the number of days the author has been active and by whether the comment contains the n-gram "thank", an important feature of the classifier. For the following analyses, we then use the \textit{equal-error threshold} over the union of the development and tests sets. At this threshold ($t=0.425$), the precision is 0.63, the recall (e.g true-positive rate) is 0.63 and the false-positive rate is 0.0034. 

\label{fig:sanity_checks}

\subsection{Understanding Attacks}

In this section, we ask a number of questions about the nature of attackers, attack timing, and moderation on Wikipedia. We investigate the answers using our machine-labeled data at the equal-error threshold. \\

%\question{How do attacks vary over time?}
%
%We took a random sample of 1M comments from 2004-2015 and computed the estimated rate of attacks within each year. \Figref{fig:prevalence_by_year} shows that the rate of attacks shot up in the early years of Wikipedia, plateaued from 2006 to 2009, and then started to decline. \\
%\begin{figure}[h]
%\centering
%\includegraphics[width=\figwidth]{./figs/prevalence_by_year.png}
%\caption{Estimated rate of attacks by year}
%\label{fig:prevalence_by_year}
%\end{figure}
% Lucas added: TODO: consider: can we add more lines to the graph, e.g. so show % of attacks by anonymous accounts, and % by accounts that contribute a lot. This would be super cool, and then maybe we can move this to end of the analysis section?

\question{What is the impact of anonymity?}

Wikipedia users can make edits either under a registered username or anonymously. In the latter case, the edits are attributed to the IP address from which they were made.\footnote{This means that, in principle, there can be multiple users editing under the same IP, and the same user editing under multiple IPs.} Table \ref{tab:anonymous_numbers} shows that, last year, 43\% of editing accounts were anonymous and these contributed 9.6\% of the comments in our dataset. 

It has been shown that anonymity provides psycho-social benefits to cyberbullies \cite{moore2012anonymity} and can lead to "heightened aggression and inappropriate behavior" \cite{ybarra2004youth}. We compare the prevalence of attacks for registered and anonymous users in Table \ref{tab:anonymous_numbers}.\footnote{A future analysis might also try to differentiate registered accounts with a long running reputation from so called sock-puppet accounts created by a user to appear as if their contributions are coming from multiple users.} This shows that the attack prevalence among comments by anonymous users is six times as high as that of registered users. Thus, while anonymous contributions are much more likely to be an attack, overall they contribute less than half of attacks. This difference of means is significant at a p < 0.0001 (t=63.8) level. These extreme values of significance are not surprising as our algorithm allows us to label data at a population level. \\

\begin{table}[h]
\centering
\noindent \begin{tabularx}{\tabwidth}{ X X X X}
  \toprule
  Anonymity & Number of Accounts & Number of Comments & Attack Prevalence \\
  \midrule
  Anonymous  &   97,742  &  191,460  & 3.1\% \\
  Registered  & 129,394  & 2,023,559 & 0.5\% \\
  \midrule
  Totals & 227,136 & 2,215,019 &  0.8\% \\
  \bottomrule
\end{tabularx}
\caption{Comment statistics by user anonymity (2015).}
\label{tab:anonymous_numbers}
\end{table}

\question{How do attacks vary with the quantity of a user's contributions?}

\begin{figure}[h]
  \centering
  \subfloat[]{\includegraphics[width=0.23\textwidth]{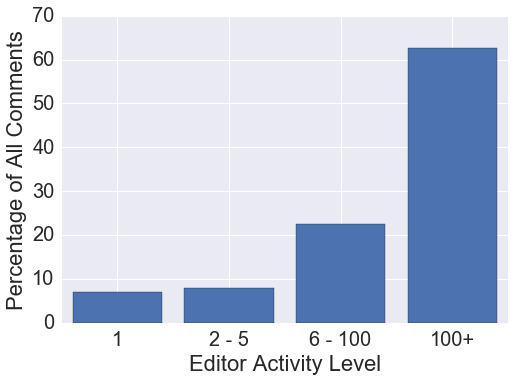}\label{fig:percentage_comments_by_activity_level}}
  \hfill
  \subfloat[]{\includegraphics[width=0.23\textwidth]{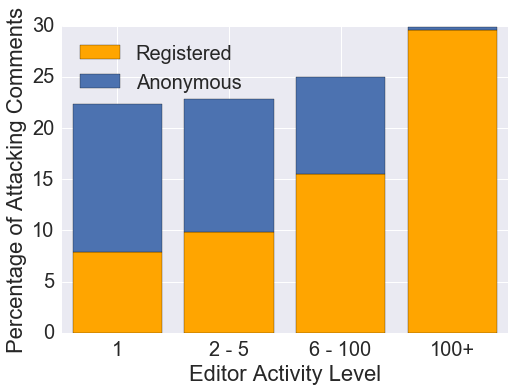}\label{fig:attacks_by_activity_level}}
  \caption{ (a). A histogram of the percentage of total comments by user activity level. (b) A histogram of the percentage of total attacks by user activity level.}
\end{figure}

Editors on Wikipedia fall along a wide spectrum in terms of their engagement with discussions on the platform. Some comment a few times a year whereas others will comment several times a week. For our purposes, a user's \textit{activity level} is the number of comments that they made in 2015. In  \figref{fig:percentage_comments_by_activity_level}, we show how many comments were made by users with different activity levels. We see that over 60\% of comments are made by users who made over 100 comments over the year. Users who made 5 or fewer comments are only responsible for 15\% of total comments.

The story completely changes when we use these same user segments to understand attacking behavior. 
\Figref{fig:attacks_by_activity_level} shows the percentage of total attacks attributable to users with different activity levels. We find that almost half of all attacks are made by users with an activity level below 5. Even controlling for the effect of anonymity that we saw earlier, more than 18\% of attacking comments are made by registered users with an activity level below 5. Users with an activity level of over 100 comments (almost all of whom are registered) are responsible for 30\% of attacking comments.
Thus, users at both low and high levels of contribution are responsible for a significant portion of attacks. 
%However, this effect is disproportionately high for those with a low level of contribution.
\\

\question{Are attacks concentrated among a few highly toxic users?}

We define the \textit{toxicity level} of a user to be the number of attacks written by that user in 2015. By segmenting users by toxicity level, we are able to uncover whether attacks are diffused among many low toxicity users or concentrated among a few users with high toxicity.

\begin{figure}[h]
  \centering
  \subfloat[]
{\includegraphics[width=0.23\textwidth]{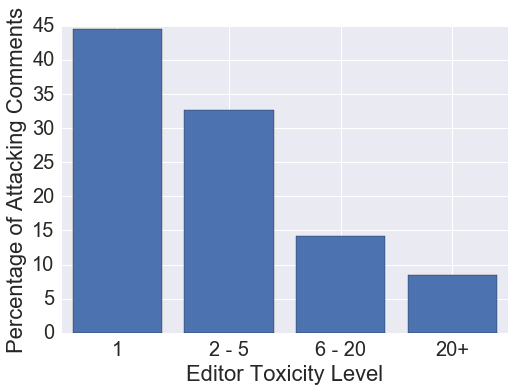}\label{fig:percentage_attacks_vs_total_attacks_made}}
  \hfill
  \subfloat[]
{\includegraphics[width=0.23\textwidth]{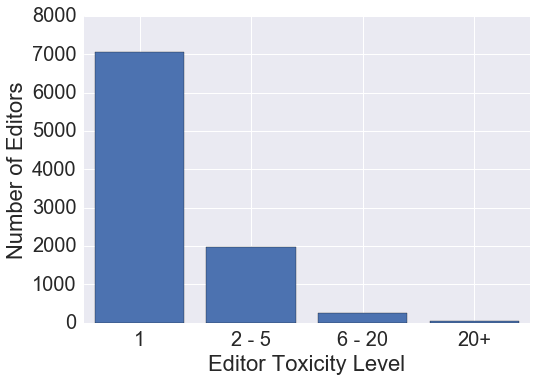}\label{fig:number_of_editors_by_toxicity}}
  \caption{ (a) A histogram of the percentage of total attacks by user toxicity level. (b) A histogram of the total number of users at each toxicity level.}
\end{figure}

\figref{fig:percentage_attacks_vs_total_attacks_made} describes the proportion of attacks made by users at different levels of toxicity. 
\figref{fig:number_of_editors_by_toxicity} provides the total number of users at each toxicity level. By comparing these figures, we see that almost 80\% of attacks come from the over 9000 users who have made fewer than 5 attacking comments. However, the 34 users with a toxicity level of more than 20 are responsible for almost 9\% of attacks. Thus, while the majority of Wikipedia's attacks are diffused infrequent attackers, significant progress could be made by moderating a relatively small number of frequent attackers. \\

%\figref{fig:percentage_comments_vs_total_attacks_made} 

\question{When do attacks result in moderation?} 

Moderators and administrators can enforce the policy on personal attacks \cite{wp:npa} by warning or blocking offending users for a period of time. 
%% The following is nice for flow, but this section is really not that big, so it feels waffly more than helpful. 
% In this section we explore how often users who make personal attacks get warned or blocked and whether attacks from moderated users are more likely to lead to future moderation events. 
%For this analysis, we increased the classification threshold to 0.8, in order to increase the precision of the classifier to 0.95. This is to ensure that 95\% of the comments labeled as personal attacks by the classifier are indeed personal attacks and are in violation of Wikipedia policy. 
Our analysis takes all attacking comments in the 2015 Wikipedia corpus and asks how many of these lead to a moderation event in the following 7 days. We find that 7.7\% of attacks are followed by a warning and 7.0\% of attacks are followed by a block within this period. 11.3\% of attacks are followed by either a warning or a block. 

As discussed in \Secref{subsec:threshold}, at the equal-error threshold, our algorithm has a precision of 0.63. After normalizing by this precision, we find that 12.2\% of the expected number of true attacks are followed by a warning, 11.1\% are followed by a block and 17.9 \% are followed by either. Thus, a high proportion of attacking comments remain unmoderated.

\begin{figure}[h]
\centering
\includegraphics[width=\columnwidth]{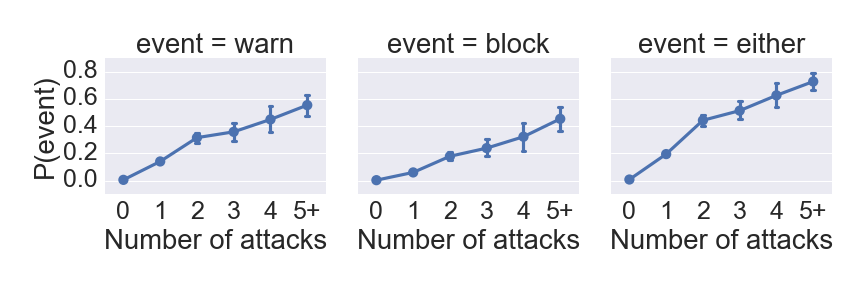}
\caption{Probability of being warned or blocked in 2015 as a function of the number of personal attacks made in 2015}
\label{fig:fraction_blocked_given_num_attacks}
\end{figure}

There are a number of factors that affect the chances of moderation, including repeated attacks and having been moderated in the past. \figref{fig:fraction_blocked_given_num_attacks} shows us that the chance of being blocked or warned increases with the number of personal attacks a user makes.
% Lucas wrote:  TODO: is this the per-comment or the per-user likelihood? And is the per-user normalized by the increased chance for writing more personal attacks? i.e. what new things exactly is this telling us?  Also: can we separate warning from blocking events?

\begin{figure}[h]
\centering
\includegraphics[width=\figwidth]{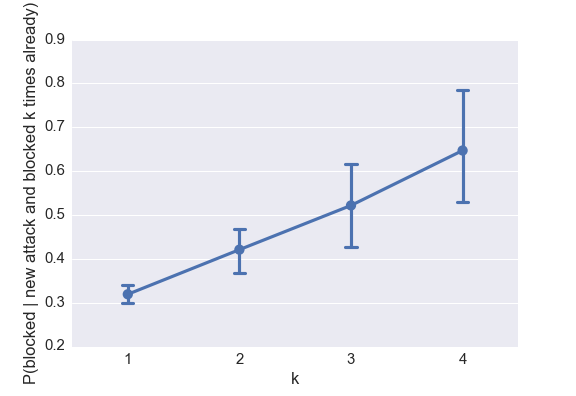}
\caption{Probability of being blocked again after a new attack as function of the number of times the user has been blocked in the past.}
\label{fig:p_of_blocked_given_new_attack_and_blocked_already}
\end{figure}

Finally, we see in \figref{fig:p_of_blocked_given_new_attack_and_blocked_already} that the likelihood of a new attack leading to a block increases with the number of times a user has been blocked in the past. This may be due to heightened scrutiny of previously blocked users. 
% Lucas wrote: TODO: Can we really not do better than leave these two possibilities? Surely we can measure if they write more attacks after they are blocked? Lets say this is future work if we don't have time to do it now.
Alternatively, it may be that blocked users make more frequent or more toxic attacks, and are hence more likely to be warned and moderated in the future. \\

\question{Is there a pattern to the timing of attacks?}

With machine labeled longitudinal data, we can generate a time-series of attacks in conversations as they occur on each page of Wikipedia. This allows us to ask whether there is a pattern to the timing of attacks in Wikipedia comments. To answer this question, we segment every comment in our corpus by whether or not it is a personal attack. We then build a neighborhood around each comment consisting of the $n$ comments that occurred on the same page immediately before and after it, excluding the central comment. For each central comment, we compute the fraction of attacks that occur in this neighborhood and call this the \textit{neighboring attack fraction} of the central comment. 

Table \ref{tab:neighboring_attack_fraction} shows the average neighboring attack fraction around attacking and non-attacking comments at different values of $n$. We see that even for small $n$, there is a significant difference (t = 56.2, p < 0.0001) in neighboring attack fractions. Indeed, the neighboring attack fraction for $n = 1$ is twenty-two times higher around attacking comments than non-attacking comments. This is a strong indication personal attacks cluster in time on Wikipedia discussions.\footnote{A follow up analysis could investigate to what extent an initial attack sparks retaliation.} It also suggests that early intervention by moderators could be an effective means of curbing the prevalence of personal attacks. 

\begin{table}[h]
\centering
\noindent \begin{tabularx}{\tabwidth}{ X X X X X }
  \toprule
  $n$ & Attacking & Non-Attacking  \\
  \midrule
  1  & 15.6 \%  & 0.7\% \\
  3  & 10.6 \% & 0.7 \%\\
  5  & 8.3 \%  & 0.8 \%\\
  \bottomrule
\end{tabularx}
\caption{Average neighboring attack fraction around attacking comments and non-attacking comments.}
\label{tab:neighboring_attack_fraction}
\end{table}

\section{Discussion \& Conclusion}
\label{sec:conclusion}
% !TEX root = 0000_main.tex

We have introduced a methodology for generating large-scale, longitudinal data on personal attacks in online discussions. After crowdsourcing the identification of personal attacks within a sample of discussion comments, machine learning classification is leveraged to scale the identification process to the whole corpus. In so doing, we explored methods for aggregating multiple human judgments per comment into training labels, compared different model architectures and text features, and introduced a technique for comparing the performance of machine learning models to human annotators. 

We illustrated our methodology by applying it to Wikipedia, generating an open dataset of over 100k high-quality human-labeled comments, 63M machine-labeled comments, and a classifier that approximates the aggregate of 3 crowd-workers. We believe this provides the largest corpus of human-labeled comments supporting the study of online toxicity to date.

By calibrating our classifier's threshold we can then perform large scale longitudinal analysis of the whole corpus of Wikipedia discussions along a wide variety of dimensions. 
We illustrate this by exploring some open questions about the nature of personal attacks on Wikipedia. 
This leads to several interesting findings: while anonymously contributed comments are 6 times more likely to be an attack, they contribute less than half of the attacks. Similarly less than half of attacks come from users with little prior participation; and perhaps surprisingly, approximately 30\% of attacks come from registered users with over 100 contributions. 
% This leads us to conclude that neither anonymous accounts, nor a small number of highlight toxic users are responsible for the majority of the personal attacks. 

These results suggest the problems associated with personal attacks do not have an easy solution.
%; removing the anonymous contributions might help, but falls far short of solving the problem. 
However, our study also shows that less than a fifth of personal attacks currently trigger any action for violating Wikipedia's policy. Moreover, personal attacks cluster in time - perhaps because one personal attacks triggers another. If so, early intervention by a moderator could have a disproportionately beneficial impact. Moreover, automated classifiers may then be a valuable tool, not only for researchers, but also for moderators on Wikipedia. They might be used to help moderators build dashboards that better visualize the health of Wikipedia conversations, or to develop systems to better triage comments for review. 

% An important points of caution should be taken with such applications: there may be unintended bias hidden with the data which becomes manifest within the model. Further research is needed to systematically understand and mitigate such biases. This should also be noted as a concern for further large scale analysis; for example if the model fails to accurately determine attacks on a group of people, it may not give accurate results when used on a sample of comments targeted at that group.

Perhaps the biggest challenge with our methodology is illustrated by our case study with Wikipedia: we used a relatively small set of annotators, 4053 in total, whom we know little about. 
% TODO: find out how many people were involved, + see histogram of their contributions for our own info.
While they have reasonable levels of inter-annotator agreement, their interpretation of a comment being a personal attack may differ from that of the Wikipedia community. Moreover, the crowdsourced data may also result in other forms of unintended bias. 

This brings up key questions for our method and more generally for applications of machine learning to analysis of comments: who defines the truth for the property in question? How much do classifiers vary depending on who is asked? 
%For example, do different demographics affect the resulting model?
What is the subsequent impact of applying a model with unintended bias to help an online community's discussion?

%% Ideally we can clarify that section, and confirm the hypothesis that users who are blocked are moderated more carefully for X time after they are blocked. Also: can we separate warning from blocking events? 
% We show that users who make more attacks are more likely to be warned or blocked and that new personal attacks are more likely to be followed by a moderation event if the user has been blocked in the past.

% This reveals that while anonymously contributed comments are 6x more likely to be an attack, they contribute less than half of the attacks; similarly accounts belonging to users with few contributions account for less than half of attacks: more than half of attacks are from accounts that have made more than 5 contributions. Thus we conclude that   This with observed temporal grouping of attacks suggests that early moderator intervention to a personal attack might be able to   

The methodology and data sets we have developed also open many other avenues for further work. The corpus of human-labeled comments can be used train more sophisticated machine learning models. It can also be used to analyze Wikipedia with traditional statistical inference, while our corpus of machine-labels can be employed to carry out further analysis that require large scale and longitudinal data. While there are many such questions to analyze, some notable examples include:

\begin{enumerate}
\item What is the impact of personal attacks on a user's future contributions?
\item What interventions can reduce the level of personal attacks on a conversation?
\item What are the triggers for personal attacks in Wikipedia comments?
\end{enumerate}

Finally, we remark that our methodology can easily be applied to different characteristics of comments, not just personal attacks. As mentioned in \secref{sec:relatedwork}, there are many taxonomies by which one can analyze the positive and negative properties of a comment. There are also many other discussion corpora to be considered.

% We will continue to work with the Wikipedia community to explore such applied outcomes of this research.

%A refined model could be used to score comments as they are posted and help moderators triage the review process. \nithum{I think this last sentence somehow implies that our model can't be used for this, which is false.}

%For example, one could measure the impact of personal attacks on user participation and retention or characterizing the conditions under which a users resort to making personal attacks.

%What the model does not get:
%see https://en.wikipedia.org/wiki/User\_talk:Jimbo\_Wales/Archive\_212\#Anti-harassment\_program

%What can you do next with the data that we publish?
%Open research that the data enables:
%Impact of attacks on retention\\
%Impact of blocks
%visualization \\
%
%What can you do next with the model?
%Bias? Mention soraya's examples (summarize her email) - comparison of gendered attacks could score low.\\
%Feature engineering work, lots that is specific to each website \\
%How classifier can be used by moderators \\

\xhdr{Acknowledgements}
% !TEX root = 0000_main.tex
{\small
The authors would like to thank
CJ Adams, Dario Taraborelli and Patrick Earley for fruitful feedback and discussions.
}

\begin{appendices}
\appendix
% !TEX root = 0000_main.tex
\section{Wikipedia Comment Corpus}
\label{appendix:corpus}

Here we describe our approach to generating a corpus of discussion comments from English Wikipedia. Every Wikipedia page, including articles and user pages, has an accompanying "talk page" that can be used for communicating with other users. Discussion pages pertaining to user pages are said to belong to the \textit{user talk namespace}, while discussions pertaining to articles belong to the  \textit{article talk namespace}. Although there are 35 talk namespaces in total, we focus on these two throughout the paper since they contain at least an order of magnitude more discussion pages and comments than the others.

\begin{figure}[h]
\centering
\includegraphics[width=\columnwidth]{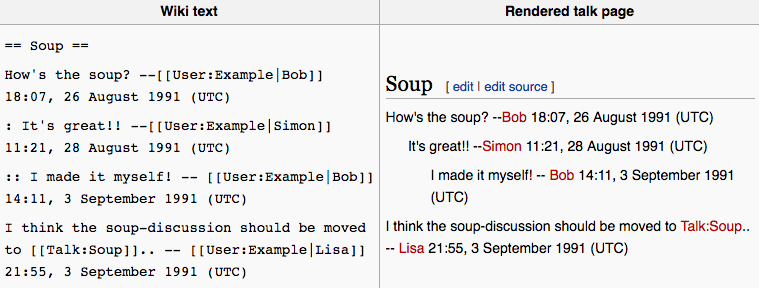}
\caption{
Example of a discussion thread taken from \cite{talk_help}. Includes the raw MediaWiki markdown or "Wiki text" and the corresponding rendering. 
}
\label{fig:discussion_example}
\end{figure}

MediaWiki, the software underlying Wikipedia, does not impose any constraints on editing talk pages. However, an edit on a talk page typically consists of a user adding a comment to a discussion in accordance with a set of formatting conventions. Figure \ref{fig:discussion_example} gives an example of a conventionally formatted discussion.

One approach to generating a corpus of comments, is to take a current snapshot of all talk pages and parse each page into discussions and comments. The downside of this approach is that comments with personal attacks are usually quickly removed and that comments on user talk pages are often removed after they have been read to reduce clutter. As a result, no single snapshot of talk pages will contain a representative or complete collection of comments made on Wikipedia.

We pursue an alternative approach, which involves processing the "revision history", which represents the history of edits on a page as a sequence of files. There is a separate file, called a revision, corresponding to the state of the article after each edit. We can compute a diff between successive revisions of a talk page to see what text was added as a result of each edit. The benefit of this approach is that it captures all content that has been added to a talk page.\footnote{Note that there is a mechanism for removing revisions from the public record, and that personal attacks are a valid reason for doing so \cite{revision_deletion}. Since this work is based entirely on publicly available data, comments introduced on deleted revisions are not included in our corpus.} The downside is that the content added during a talk page edit is not always a full, new comment. The content added can also represent a modification of an existing comment. For completeness, we include the text added in these types of edits in our corpus. %Figure \ref{fig:diff_example} illustrates a clean example of a diff computed over successive revisions of an example talk page.

%\begin{figure}[h]
%\centering
%\includegraphics[width=\figwidth]{./figs/diff_example.png}
%\caption{
%Example visualization of a diff representing the addition of a comment to a discussion. \ellery{This can go if we need space.}
%}
%\label{fig:diff_example}
%\end{figure}

In practice, we processed the revision history from a public dump of English Wikipedia made available on 2016-01-13. To generate the set of diffs from the revision history, we used the existing \textit{mwdiffs}~\cite{mwdiffs} python package along with the standard longest-common-substring diff algorithm. For the purpose of this study we define a talk page comment as the concatenation of the MediaWiki markup added during an edit of a talk page. We also compute a clean, plain-text version of each comment by stripping out any html or MediaWiki markup, which we use in the crowd-sourcing task described below. 

While manually inspecting the data, we found that a large portion of comments left on talk pages (20\%-50\%, depending on the namespace) were clearly administrative in nature and generated using a bot or template. In this study we are interested in comments made by human users in the context of discussions. 
%We exclude all messages from a small number of users whose user-names match a set of regular expressions. 
After using a regular expression to filter out all messages from these users, we still observed a large number of administrative comments with little or nor modification on many user talk pages. We generated a another set of regular expressions to remove the most commonly occurring comments of this nature. Table \ref{tab:comment_corpus} gives the number of comments in the user and article talk namespaces after each filtering step.

\begin{table}[h]
\centering
\noindent \begin{tabularx}{\tabwidth}{ X X X X}
  \toprule
  Namespace & All & No Bot & No Bot/Admin \\
  \midrule
  User  &   47.3M  &  36.5M &  24.2M\\
  Article  & 47.8M  & 39.2M & 39.2M\\
  \midrule
  \midrule
  Totals & 95.1M & 75.7 & 63.4M    \\
  \bottomrule
\end{tabularx}
\caption{Summary statistics of comment corpus broken down by namespace. We first filter out all comments from bot accounts and then filter out messages containing templates used for administrative purposes.  
}
\label{tab:comment_corpus}
\end{table}

\end{appendices}

\bibliographystyle{abbrv}
\bibliography{bibliography}

\end{document}